\documentclass[sigplan,10pt]{acmart}
\startPage{1}

\setcopyright{none}

\usepackage{booktabs}   
\usepackage{subcaption} 
\usepackage[normalem]{ulem}
\usepackage{makecell}
\usepackage[ruled,vlined,linesnumbered]{algorithm2e}
\usepackage{pifont}
\usepackage{amsfonts}
\usepackage{multirow}

\begin{document}

\title{RLHFSpec: Breaking the Efficiency Bottleneck in RLHF Training via Adaptive Drafting}

\author{Siqi Wang, Hailong Yang$^\ddagger$, Junjie Zhu, Xuezhu Wang, Yufan Xu, Depei Qian}

\thanks{$^\ddagger$Corresponding author}





\fancyhead{}  
\renewcommand\footnotetextcopyrightpermission[1]{} 

\begin{abstract}
Reinforcement Learning from Human Feedback (RLHF) is an important fine-tuning technique for large language models (LLMs) and comprises three stages: generation, inference, and training. The generation stage generates samples that are then used to infer learnable experiences for training. We observe that the generation stage is the bottleneck of the entire execution process and consider it a key point for optimization. Specifically, we realize the first attempt to integrate speculative decoding into the RLHF generation stage 
and propose \textit{RLHFSpec}, an RLHF system that accelerates generation execution with efficient speculative decoding and sample reallocation. 
To fully exploit the performance potential provided by speculative decoding, especially dealing with the dynamic workload of the generation stage, \textit{RLHFSpec} proposes a workload-aware drafting strategy selection mechanism, which selects the near-optimal strategy by jointly considering the verification cost and the number of accepted tokens. Moreover, \textit{RLHFSpec} also proposes sample reallocation to fully utilize the GPU resources, and optimizes it with an efficient sample migration mechanism. The experimental results show that the \textit{RLHFSpec} can achieve higher throughput in the generation stage compared to state-of-the-art works. Moreover, due to the effective alleviation of the generation bottleneck, \textit{RLHFSpec} also shows significant performance speedup in the entire RLHF execution. 
\end{abstract}

\maketitle

\section{Introduction}

Recently, large language models (LLMs) such as GPT~\cite{achiam2023gpt}, Llama~\cite{grattafiori2024llama}, and Deepseek~\cite{guo2025deepseek} have achieved significant success in various domains within both industry and academia~\cite{adiwardana2020towards, acroller2020recipes, acnallapati2016abstractive, acpaulus2017deep, acsee2017get, acchen2021evaluating, acxu2015show, acyang2016review}. Reinforcement Learning from Human Feedback (RLHF) is a significant fine-tuning technique for LLMs that has facilitated the evolution of their language understanding capabilities, playing a critical role in the practical applications of LLMs~\cite{bai2022training,ouyang2022training}.

The RLHF training process can be divided into three stages: \textit{generation}, \textit{inference}, and \textit{training}. The \textit{generation} stage generates responses for samples using autoregressive decoding, including a prefill step and multiple decoding steps. The decoding time is proportional to the response length, significantly increasing the generation time. The \textit{inference} stage then utilizes responses generated from the previous stage to produce learnable experiences through a forward pass. In the \textit{training} stage, generated responses and experiences are employed to update models through both a forward pass and a backward pass. Notably, we observe that the generation stage dominates the overall time in RLHF execution due to its multiple decoding steps (details in Section~\ref{subsec:tail}). 

Moreover, due to the low parallelism inherent in the decoding steps to generate responses autoregressively, generation stage is unable to fully utilize all computational resources. Even worse, with the recent emergence of Chain of Thought (CoT), the LLMs generate longer responses with the thinking and reasoning enabled. In such a case, the length of the response varies depending on the user inputs, resulting in a long-tailed distribution phenomenon (details in Section~\ref{subsec:tail}). 
In the context of RLHF, shorter samples gradually terminate their decoding, leaving only a few long-tailed samples in the system. This results in GPUs experiencing prolonged periods of processing only a limited number of samples, which leads to lower GPU utilization and inefficient execution. Thus, to enhance the performance of RLHF training, addressing the above limitation in the generation stage is crucial.

Speculative decoding shows superior performance and efficiency when deployed in online serving~\cite{miao2024specinfer,butler2024pipeinfer}, addressing the challenges of low parallelism and enhancing resource utilization. As the generation stage exhibits the same autoregression computation pattern as online serving, adopting speculative decoding has the potential to alleviate the performance bottleneck in RLHF generation. However, from the workload perspective, unlike online serving, RLHF generation resembles offline inference, where the total number of processing samples is fixed. Moreover, the optimization goal is also different from online serving, which is to complete all samples in the shortest time, prioritizing throughput over latency. Naively integrating speculative decoding into RLHF generation cannot fully exploit the performance potential due to the static drafting strategy previously used in online serving (details in Section~\ref{subsec:spec}).

Specifically, in RLHF's generation stage, the samples are gradually processed, leading to a \textbf{dynamic workload}.
During the initial phase of generation, resource utilization is relatively high as responses are generated for all samples. As shorter samples finish, the resource utilization of corresponding generation instances drops. Prior studies of speculative decoding typically select a drafting strategy empirically without adjusting it during runtime, overlooking the impact of workload changes on optimal strategy selection. For instance, always adopting an aggressive strategy can incur substantial verification cost in the initial phase, potentially offsetting the benefits of speculative decoding. Conversely, recklessly using a conservative strategy will leave precious resources underutilized in the later phase, preventing speculative decoding from achieving higher performance speedup (details in Section~\ref{subsec:naive}). Therefore, we derive \textbf{Challenge 1: Static drafting strategy is sub-optimal for RLHF generation with dynamic workloads.} To address this challenge, we propose a workload-aware drafting strategy selection mechanism when integrating speculative decoding into RLHF generation, which can quickly identify the near-optimal drafting strategy for a specific workload.
Moreover, due to the autoregressive nature of generation, the sample response length is inherently unpredictable. In online serving, continuous batching~\cite{yu2022orca} with iteration-level scheduling mitigates the varying response length by balancing workloads across instances and maintaining high utilization of hardware. In contrast, RLHF generation processes a fixed number of samples, preventing the batching technique from being effective. Therefore, certain generation instances may remain fully loaded when assigned long-tailed samples, whereas others quickly become idle due to shorter samples assigned, creating \textbf{severe load imbalance}. Existing speculative decoding methods~\cite{butler2024pipeinfer,cai2024medusa,chen2023accelerating} statically allocate samples to instances, exacerbating the resource underutilization problem. Consequently, even though individual instances may achieve near-optimal throughput under a workload-aware drafting strategy, the overall system throughput remains far from optimal (details in Section~\ref{subsec:sampleallocation}). Therefore, we derive \textbf{Challenge 2: Fixed sample allocation results in GPU resource underutilization.} We find that reallocating samples across instances effectively addresses this issue and improves the use of idle resources. We propose an efficient reallocation policy that improves system throughput, together with a lightweight migration mechanism that overlaps computation and communication to enable effective reallocation.

This paper realizes the first attempt to integrate speculative decoding into the RLHF system, which novelly addresses the performance bottleneck and low resource utilization during the generation stage. Specifically, this paper makes the following contributions:
\begin{itemize}
    \item We develop a workload-aware drafting strategy selection mechanism to select the near-optimal strategy for dynamic generation workloads by jointly considering the verification cost and the number of accepted tokens. 
    \item We propose sample reallocation as a key metric to fully utilize GPU resources and enhance system throughput. We introduce an efficient policy to determine the reallocation strategy and realize it with a two-stage sample migration mechanism.
    \item We develop an RLHF system \textit{RLHFSpec} that accelerates RLHF generation with efficient speculative decoding and sample reallocation. We perform a thorough experimental comparison with existing systems and demonstrate significant performance improvements over the state-of-the-art works. 
\end{itemize}
\section{Background}

\subsection{RLHF Workflow}
The workflow of one iteration in RLHF is composed of three stages (i.e., generation, inference, and training) and four models (i.e., actor model, reference model, reward model, and critic model). The \textit{generation stage} is an autoregressive decoding process composed of multiple forward passes, during which the actor model generates responses based on given prompts. Due to the autoregressive nature, the generation is performed iteratively, which decodes the next token based on the preceding token sequence in each step until reaching a termination token. As the lengths of the responses are variable, the samples in this stage will not be completed at the same time but rather terminate progressively. The \textit{inference stage} is a forward pass over the combination of prompts and generated responses. The reward model, critic model, and reference model use the prompts and responses to create the learnable experiences. The \textit{training stage} is an supervised training iteration, composed of a single forward pass and a backward pass. The actor model and critic model are updated based on the generated responses and the learnable experiences. The next RLHF iteration then utilizes these updated models for generation and inference. 
Among these stages, the \textit{generation stage}, which necessitates multiple decoding steps, accounts for the majority of the time spent on RLHF execution (more than 68.4\%)~\cite{zhong2025optimizing,mei2024realhf} (details in Section~\ref{subsec:tail}).

\subsection{Speculative Decoding}
\label{subsec:spec}
The fundamental idea of speculative decoding is to use a small draft model (SSM) to generate speculative tokens autoregressively (a.k.a., draft generation). These speculative tokens are then fed into the LLM to be verified in a single decoding step (a.k.a., LLM verification). If the speculative tokens from the SSM are consistent with LLM, the LLM accepts these tokens. Since actual draft generation is carried out with the SSM, which is an order of magnitude faster than LLM, it can effectively accelerate the LLM serving. The more tokens generated by the SSM that are accepted by the LLM, the greater the acceleration of LLM serving will be. Moreover, several works~\cite{chen2023accelerating,leviathan2023fast} have proved that speculative decoding is consistent with the distribution of autoregressive decoding, and it has no degradation of inference precision.

\begin{figure}[htbp]
    \centering
    \includegraphics[width=\columnwidth,trim={0cm, 0.5cm ,0cm, 0.5cm}]{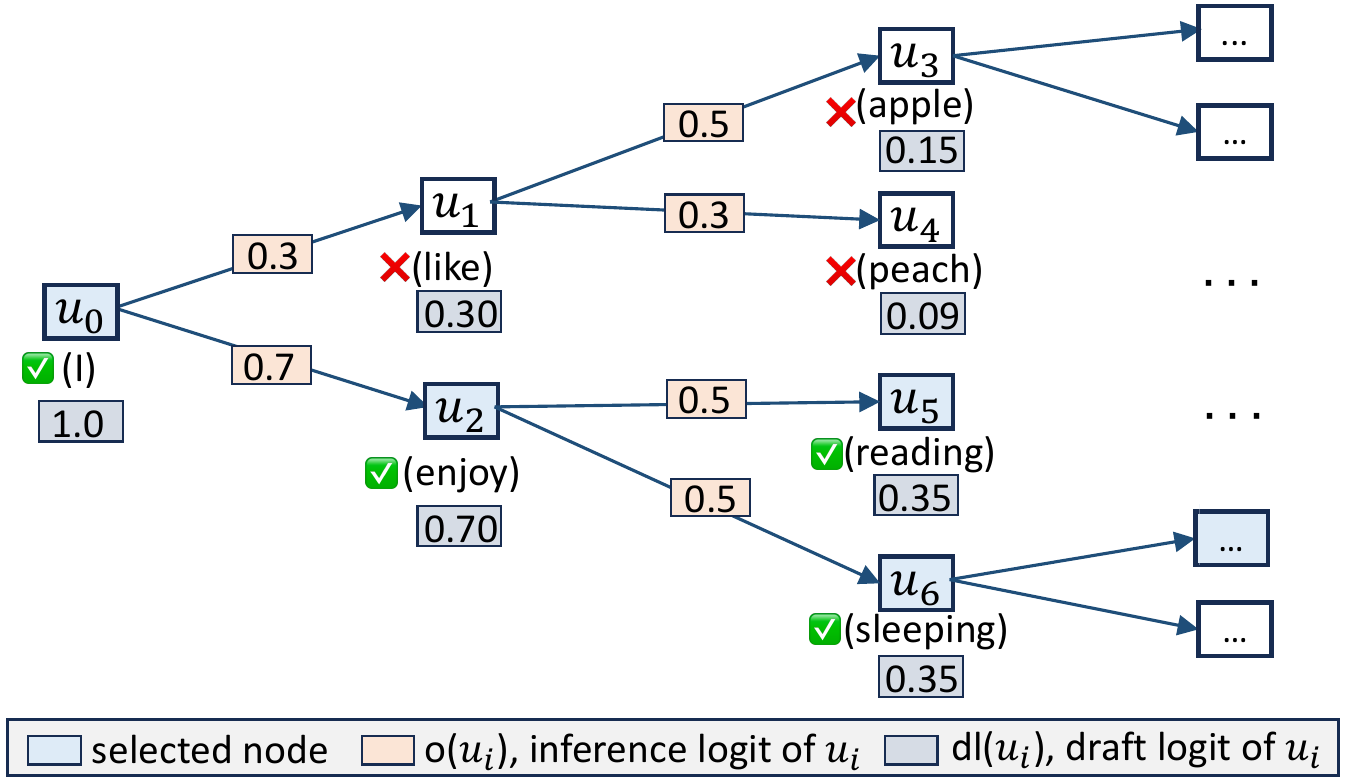}
    \caption{Illustration of tree-based speculative decoding.}
    \label{fig:spec}
\end{figure}

To enhance the acceptance rate of speculative tokens, recent works have proposed tree-based speculative decoding~\cite{li2024eagle,miao2024specinfer}. As shown in Figure~\ref{fig:spec}, during the draft generation process, multiple draft tokens are selected at each step to form a tree, where each node represents a token and each branch represents a token sequence awaiting verification (e.g., \textit{"I enjoy sleeping"}). In the following text, we use “node” and “token” interchangeably. Since not all tokens possess a high acceptance probability, verifying all tokens in the tree will result in unnecessarily high verification cost. To solve this issue, recent works define the \textit{draft logit} ($dl(u)$) of the node $u$ as the product of the SSM inference logits of all nodes on the path from the root node to $u$, namely $dl(u) = \prod_{v\in Path(root, u)} o(v)$, where $o(v)$ denotes the inference logit of a node $v$. The top $n$ nodes with the highest $dl$ are selected to form a connected tree for verification, which significantly improves the performance. For example, in Figure~\ref{fig:spec}, $dl(u_6) = o(u_0)\times o(u_2)  = 0.7 \times 0.5 = 0.35$. With $n=4$, $u_0$,$u_2$,$u_5$, and $u_6$ are top-4 nodes with highest $dl$. Thereby, the token set $\{"I", "enjoy", "reading", "sleeping"\}$ is sent to LLM for verification. It is noteworthy that the $n$ is a user-specified and fixed parameter, referred to as the \textit{draft token num}, and the selection of $n$ is referred to as the \textbf{\textit{drafting strategy}}.

\section{Motivation}
\subsection{Long-tail Phenomenon in Generation Stage}
\label{subsec:tail}
\begin{figure}[htbp]
    \centering
    \begin{minipage}{0.56\linewidth}
        \centering
        \includegraphics[width=\columnwidth,trim={0cm 1.2cm 0cm 0.8cm}]{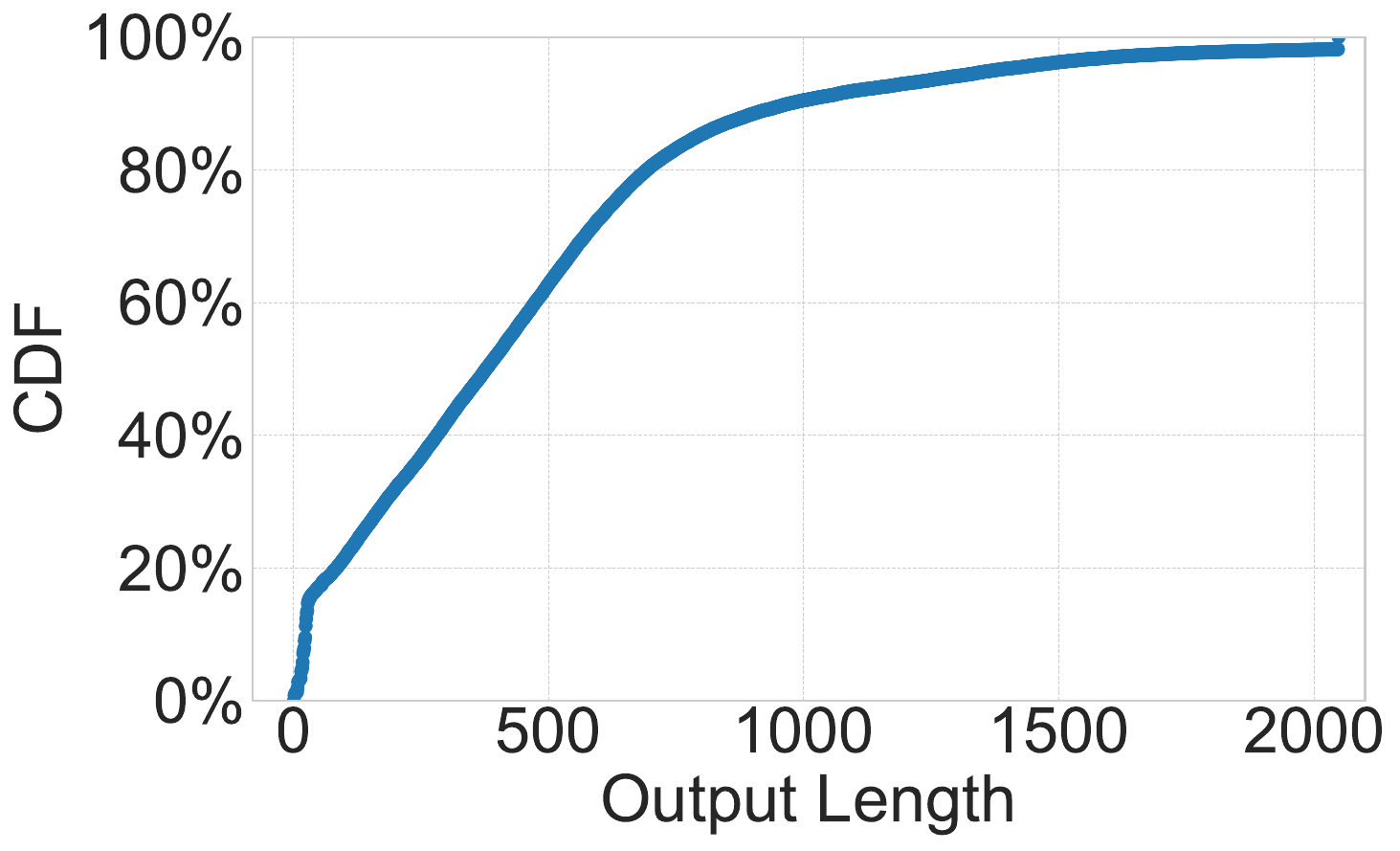}
        \caption{The CDF of generation output length.}
        \label{fig:cdf}
    \end{minipage}\hfill
    \begin{minipage}{0.43\linewidth}
        \centering
        \includegraphics[width=\columnwidth,trim={0cm 1.1cm 0cm 0.69cm}]{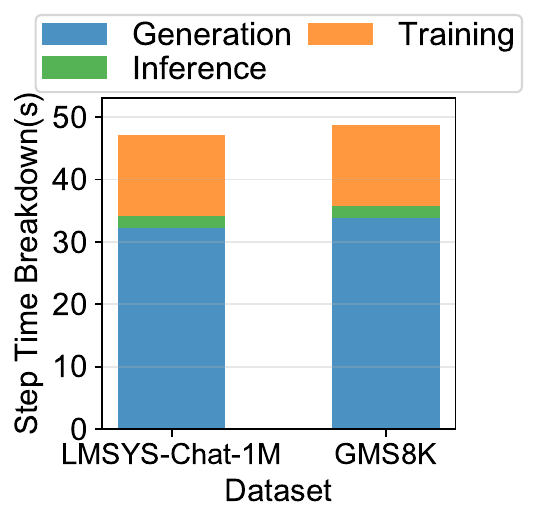}
        \caption{The RLHF iteration breakdown. \\}
        \label{fig:stepbreak}
    \end{minipage}
\end{figure}

We observe that the response length of generated samples exhibits a long-tailed distribution. Figure~\ref{fig:cdf} shows the CDF of the output length distribution of LMSYS-Chat-1M dataset~\cite{zheng2023lmsyschat1m}, which contains one million real-world conversations with 25 state-of-the-art LLMs. As shown in Figure~\ref{fig:cdf}, the median length is only 378, while the 95th percentile is 1373, which is nearly four times the median. This results in a continuous reduction of the load during generation, and gradually, as the remaining samples decrease, the GPU resources can not be fully utilized, leading to inefficient execution. Moreover, as shown in Figure~\ref{fig:stepbreak}, the generation stage occupies a significant portion of time during the RLHF training process, exceeding 68.4\% of the total execution time. Therefore, optimizing the inefficient execution of the generation stage is crucial for improving the overall performance of RLHF training.

\subsection{Static Drafting strategy Leads to Sub-optimal Performance}  
\label{subsec:naive}
\begin{figure}[htbp]
    \centering
    \includegraphics[width=\columnwidth,trim={0cm 0cm 0cm 0cm}]{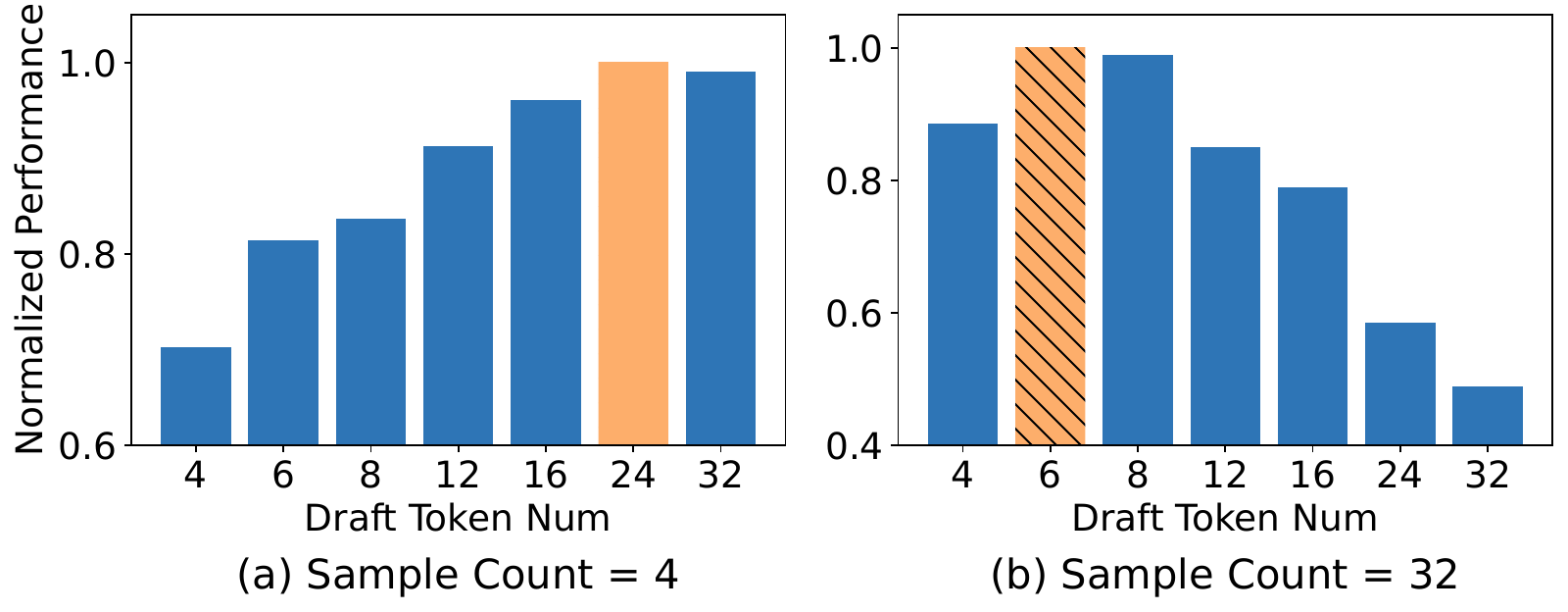}
    \caption{Throughputs using speculative decoding under different drafting strategies (i.e., draft token num) and sample counts. Reported throughputs are normalized by the highest throughput across all drafting strategies.}
    \label{fig:draft_num}
\end{figure}

As discussed in Section~\ref{subsec:spec}, existing speculative decoding works~\cite{li2024eagle,miao2024specinfer} employ a static drafting strategy (i.e., fixed \textit{draft token num}). However, a naive integration of the speculative decoding approach to the generation stage cannot fully exploit the performance opportunities. Figure~\ref{fig:draft_num} illustrates the performance using varying \textit{draft token num} ($n$) under different workloads. As shown in Figure~\ref{fig:draft_num}, there are performance disparities among different drafting strategies, and the optimal strategy varies with different generation workloads. 
During the early stage of generation, where the workload is relatively high (Figure~\ref{fig:draft_num}(b)), employing strategies with a relatively high value (e.g., $n=24$) incurs a substantial workload and high verification cost, leading to reduced acceleration. In the latter stage, where the workload diminishes (Figure~\ref{fig:draft_num}(a)), the reduced verification pressure enables a higher $n$ to yield a larger number of accepted tokens and consequently result in enhanced performance. Conversely, when using a smaller value (e.g., $n=6$), although the early stage can achieve a higher acceleration rate compared to the case where $n=24$, the acceleration rate remains relatively low in the latter stage due to the smaller number of accepted tokens.
Experimental results show that the dynamic nature of the generation workload prevents a static drafting strategy from consistently achieving optimal performance. To fully leverage the benefits of speculative decoding, the drafting strategy must be adapted dynamically to the workload.
\textbf{Observation 1: Determining the optimal drafting strategy for different workloads enhances generation performance by balancing the verification cost and the number of accepted tokens.}

\subsection{Fixed Sample Allocation Hinders Generation Performance}
\label{subsec:sample_motivation}
\begin{figure}[htbp]
    \centering
    \includegraphics[width=0.95\columnwidth,trim={0cm 0cm 0cm 0cm}]{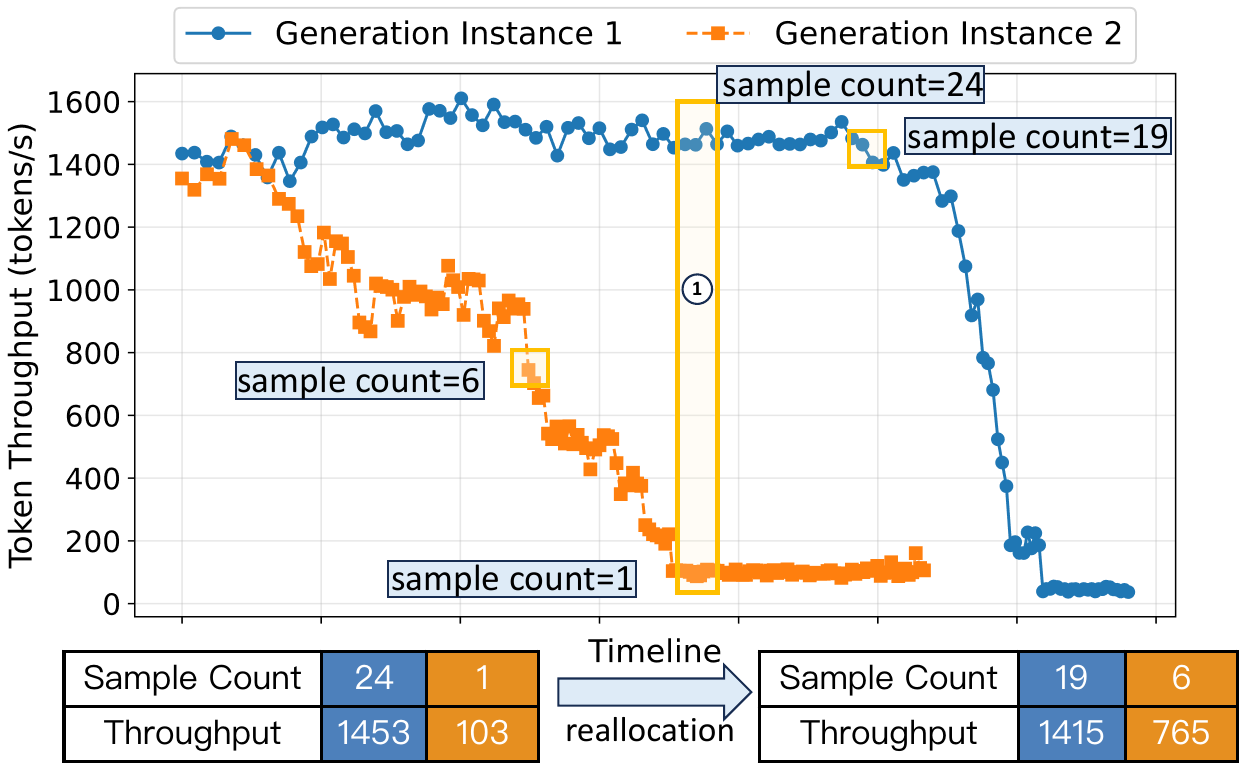}
    \caption{Throughput variation curves of generation instances over time and optimization opportunities.}
    \label{fig:throughput_motivation}
\end{figure}

Figure~\ref{fig:throughput_motivation} illustrates the throughput variation curves of two generation instances. As shown in Figure~\ref{fig:throughput_motivation}, generation instance 1 (denoted as \textit{ins.1}) handles a greater number of long-tailed samples, maintaining a high load and throughput for the majority of the execution time. In contrast, generation instance 2 (denoted as \textit{ins.2}) processes more short samples, resulting in a continuous decline in throughput as the load decreases. At time slot \ding{172}, \textit{ins.1} processes 24 samples with a throughput of 1,453 tokens/s, whereas \textit{ins.2} processes only 1 sample with a throughput of merely 103 tokens/s. At this moment, the total throughput of these instances is 1,556 tokens/s. If we do not change the total number of samples but rather reallocate samples by moving 5 samples from \textit{ins.1} to \textit{ins.2}, adjusting the sample distribution from (24+1) to (19+6). The throughput of \textit{ins.1} remains relatively stable at 1,415 tokens/s, while the throughput of \textit{ins.2} significantly increases to 765 tokens/s. This reallocation results in a substantial improvement in total throughput, increasing it to 2,180 tokens/s. \textbf{Observation 2: Reallocating samples in RLHF generation reveals promising opportunities to enhance system throughput.}

\label{subsec:sampleallocation}


\section{Design Overview}
\begin{figure}[htbp]
    \centering
    \includegraphics[width=0.85\columnwidth,trim={0cm 0.3cm 0cm 0.1cm}]{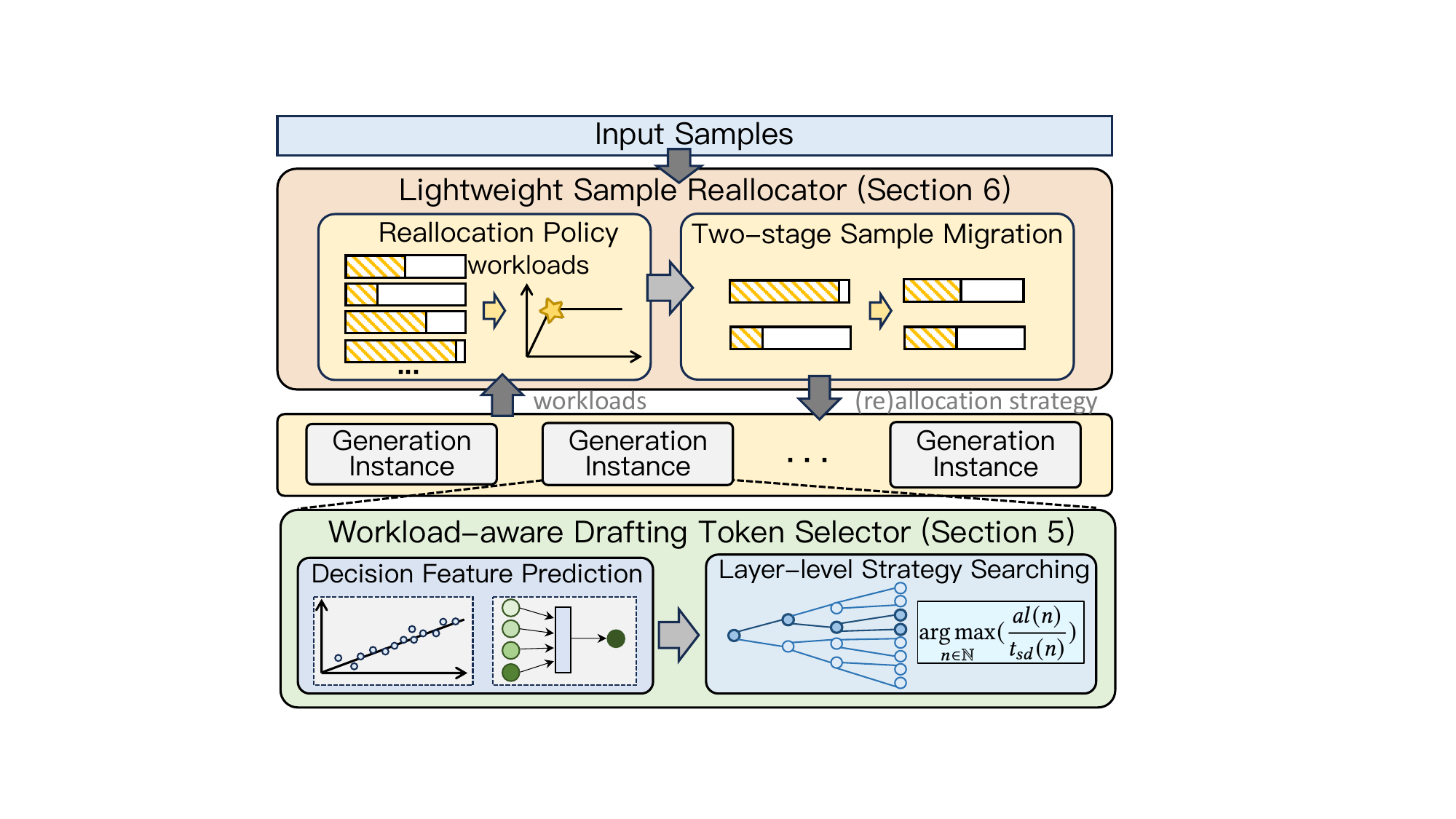}
    \caption{Design overview of \textit{RLHFSpec}.}
    \label{fig:design_overview}
\end{figure}

To leverage the observations discussed above, we propose \textit{RLHFSpec}, an RLHF system that accelerates RLHF generation with efficient speculative decoding and sample reallocation. The \textit{RLHFSpec} system is composed of two major components, including workload-aware drafting strategy selector (Section~\ref{sec:workload}) and lightweight sample reallocator (Section~\ref{sec:allocation}). The workflow of the \textit{RLHFSpec} system is shown in Figure~\ref{fig:design_overview}.

Training samples are first sequentially allocated to the generation instances, and the instance workloads are reported by the instances periodically. To achieve high throughput, \textit{RLHFSpec} introduces 
the \textit{lightweight sample reallocator}. The reallocator monitors instance workloads. When inefficient execution is detected, the reallocation is triggered. An efficient reallocation policy is utilized to generate the strategy, and a two-stage sample migration mechanism is proposed to reduce reallocation overhead. The reallocation strategy is then sent to the corresponding generation instances.

Upon receiving the allocation/reallocation strategy, to ensure that generation instances can fully leverage the performance benefits provided by speculative decoding under dynamic workloads, \textit{RLHFSpec} also introduces the \textit{workload-aware drafting strategy selector}. For each speculative step of every instance, the \textit{workload-aware drafting strategy selector} is employed to identify the near-optimal drafting strategy for the current workload. This approach enhances performance by jointly considering the number of accepted tokens and the verification costs associated with various workloads.

\section{Workload-aware Drafting Strategy Selection}
\label{sec:workload}
To address the inefficiencies discussed in Section~\ref{subsec:naive}, we explore the adaptive selection of near-optimal drafting strategies based on workload. As the load continuously changes during execution, our decision needs to be made at each step. Thereby, it is essential to implement lightweight methodologies to ensure that the decision-making process remains cost-effective and does not negatively impact performance. 


\subsection{Optimization Objectives}
To select the near-optimal drafting strategy, we first analyze how different drafting strategies impact generation performance. For a specific workload, we define the effective drafting strategy as one that achieves the highest speedup compared to autoregressive decoding. Specifically, for one-step speculative decoding, the execution time is denoted as $t_{sd}(n)$, which is determined by the \textit{draft token num} $n$. The number of accepted tokens is denoted as $al(n)$, which is also influenced by $n$. Assuming the execution time for one-step autoregressive decoding is $t_{ar}$. To generate tokens equivalent to those produced by speculative decoding, $al(n)$ executions are required. Therefore, compared to autoregressive decoding, the speedup of speculative decoding is formulated as: 


\begin{equation}
    \begin{split}
    \footnotesize
    Speedup(n)\ =\ \frac{al(n) \times t_{ar}}{t_{sd}(n)}
    \end{split}
    \label{eq:speedup}
\end{equation}

To optimize the speculative performance under various workloads, the key factor is choosing the optimal value for $n$ to maximize the speedup. We define the optimization function as Equation~\ref{eq:optimize}, where $t_{ar}$ depends only on the samples regardless of $n$. To obtain the decision features $al(n)$ and $t_{sd}(n)$ under specific $n$ without actual execution, we propose two lightweight prediction methods.

\begin{equation}
    \begin{split}
    n_{optimal} &= \mathop{\arg\max}\limits_{n \in \mathbb{N}}(\frac{al(n) \times t_{ar}}{t_{sd}(n)}) =\mathop{\arg\max}\limits_{n \in \mathbb{N}}(\frac{al(n)}{t_{sd}(n)})
    \end{split}
    \label{eq:optimize}
\end{equation}

\subsection{Decision Features Prediction}
\label{subsec:prediction}
\begin{figure}[htbp]
    \centering
    \includegraphics[width=\columnwidth, trim={0cm 0cm 0cm 0cm}]{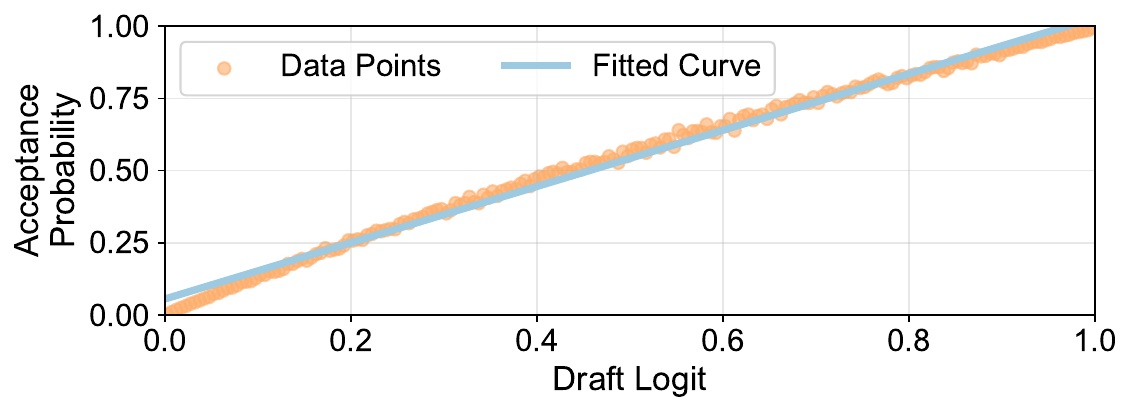}
    \caption{The fitted curve of the draft logit and acceptance probability.}
    \label{fig:acc_pred}
\end{figure}

\textbf{The number of accepted tokens} $al$ is hardly determined beforehand, as it depends on the LLM to verify the speculative tokens. \textit{RLHFSpec} addresses this challenge by exploiting a key characteristic of speculative decoding: the draft logits of the SSM have a positive correlation relationship to the probability that each speculative token can be accepted by the LLM (i.e., acceptance probability). This is due to the fact that the SSM is typically distilled from the LLM~\cite{zhou2023distillspec,li2024eagle}, ensuring that the logits of the SSM closely align with those of the LLM. Our experimental results in Figure~\ref{fig:acc_pred} further demonstrate this point, which shows a significant linear correlation trend. Based on this observation, we fit a function (i.e., $F: X->Y$) between draft logits and token acceptance probability based on offline profiling data. The profiling overhead is negligible compared to the RLHF training (details in Section~\ref{subsec:overhead}). Moreover, we also collect online data to update the function to ensure accuracy. 

\begin{figure}[htbp]
    \centering
    \includegraphics[width=0.95\columnwidth]{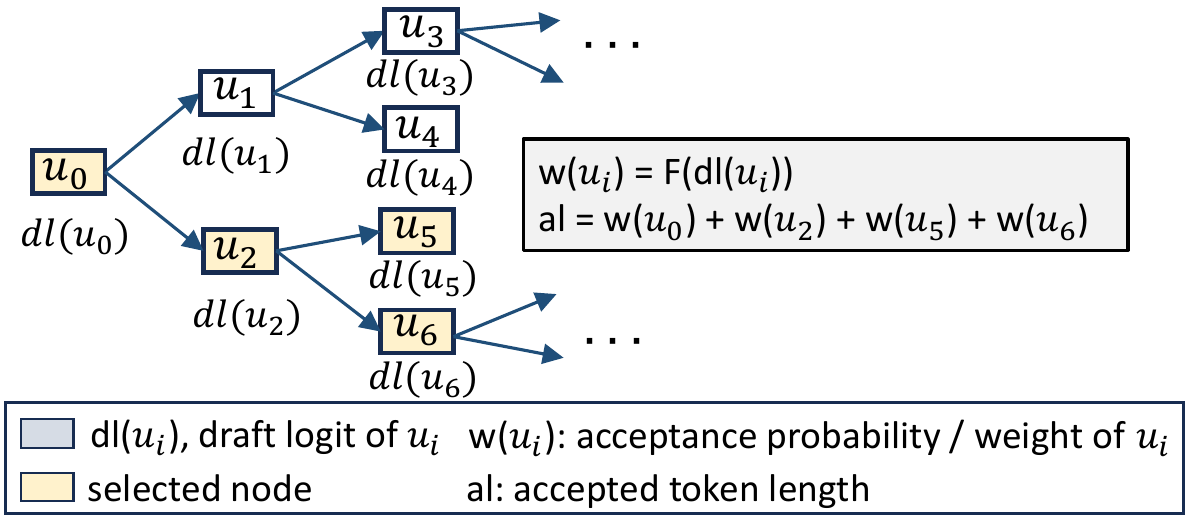}
    \caption{Illustration of $al$ calculation process.}
    \label{fig:tree}
\end{figure}

Based on the acceptance probability prediction, for a given speculative tree, we can obtain the acceptance probability for any node $u$ (i.e., speculative token) within the tree, denoted as node weight $w(u)$. 
For a set of selected nodes, the prediction result for the $al$ can be obtained by summing the weight of each selected node, namely $al = \sum_{u \in selected\ set}w(u)$. Figure~\ref{fig:tree} illustrates an example to obtain the $al$ with the \textit{draft token num=4}. We first obtain the weights for node $u_i$ using its draft logit $dl(u_i)$ through the fitting function $F$, represented as $w(u_i) = F(dl(u_i))$. After that, we sum the weights of the selected nodes to obtain $al$. As shown in Figure~\ref{fig:tree}, $al = w(u_0)+w(u_2)+w(u_5)+w(u_6)$.


\textbf{The execution time} ($t_{sd}$) is predicted as follows. The speculative decoding is composed of draft generation followed by LLM verification. Given that the overhead of draft generation is associated with the establishment of the speculative tree, it remains invariant regardless of the selected drafting strategy and can be regarded as a constant. For LLM verification cost, we observe a strong correlation with the following features. The LLM verification primarily comprises two time-consuming computations: \textit{attention} and \textit{feedforward network (FFN)} calculations. The \textit{attention} primarily incurs cost due to KVCache loading, which correlates with the size of the KVCache for tokens that have already been computed. This cost is specifically related to the cumulative sequence lengths of all samples within the batch, denoted as $N_{seq}$. The cost of the \textit{FFN} arises from matrix multiplication computations, with the computational intensity related to the sum of the \textit{draft token num} across all samples in the batch ($N_{draft}$). Moreover, hardware configurations, including bandwidth and computational capability, also impact the verification cost. Based on these features, we construct a regression model and perform offline profiling of the data for model training. The profiling overhead is minimal and can be considered negligible compared to the RLHF training (details in Section~\ref{subsec:overhead}).

Moreover, to achieve lightweight prediction, we propose further optimization. We find that variations in $N_{seq}$ and $N_{draft}$ within a range do not affect the final $t_{sd}$ and propose a bucket-based caching mechanism. We classify the value ranges of $N_{seq}$ and $N_{draft}$ into multiple buckets, and assume that the combinations of $N_{seq}$ and $N_{draft}$ within each bucket exhibit the same $t_{sd}$. We hold a cache to store the predictions. Specifically, whenever a $N_{seq}$ and $N_{draft}$ pair is encountered, the predictor first searches in cache. If a prediction result for the input within the same bucket already exists in the cache, it is directly retrieved as a cache hit. If the cache miss occurs, the predictor can derive the result in less than a millisecond.

\subsection{Layer-level Strategy Searching}
Our goal is to select the \textit{draft token num} $n$ to achieve high speculative performance. We iterate through different values of $n$ to obtain the node set (i.e., token set) $S(n) = \{ u_{i_1}, u_{i_2}, \ldots, u_{i_n} \}$, which, similar to previous work, consists of the top $n$ weighted nodes. For each $n$, we calculate the optimization objective value in Equation~\ref{eq:optimize} and select the strategy with the highest value as the final strategy. We first propose two principles to guide our selection: (1) $S(n)$ is selected from the first $n$ layers of the tree, as the inclusion of additional layers hinders $S(n)$ from forming a connected tree, leading to an inefficient selection; (2) $S(n+1) = S(n) \cup \{ u_{max} \}$, where $u_{max}$ is the node with the maximum weight among the unselected nodes.

Based on the above principles, we employ layer-level searching to traverse the speculative tree. Upon reaching layer $m$, we add the nodes at this layer to the max priority queue, and retrieve $u_{max}$ from the max priority queue to form $S(m)$. We then utilize the predictors to obtain the $al(m)$ and $t_{sd}(m)$, and obtain the optimization objective value in Equation~\ref{eq:optimize}. The strategy with the highest value is selected as the final decision.

We further implement pruning to accelerate the searching process. We observe that the maximum optimization objective value in Equation~\ref{eq:optimize} occurs before it begins to decline. Specifically, we define the $al$ and $t_{sd}$ of current and next search step as $al(n)$/$t_{sd}(n)$ and $al(n+1)$/$t_{sd}(n+1)$, respectively. The increments of $al$ and $t_{sd}$ between these two steps are denoted as $\Delta al$ and $\Delta t_{sd}$. According to principle 2, $\Delta al$ is the maximum node weight among the unselected nodes, which exhibits a downward trend. For $\Delta t_{sd}$, as the number of verified tokens increases, the hardware capacity gradually becomes saturated, resulting in an upward trend in $\Delta t_{sd}$. Thus, $\frac{\Delta al}{\Delta t_{sd}}$ decreases as the search progresses. According to the sugar water inequality (Equation~\ref{eq:sugar}), when $\frac{\Delta al}{\Delta t_{sd}}$ decreases to less than $\frac{\Delta al(n)}{\Delta t_{sd}(n)}$, further increases in $n$ will result in diminished gains, thereby permitting the termination of the search. In \textit{RLHFSpec}, when the search process detects a continuous decrease in the optimization objective value, we will implement an early stopping for the search.
\begin{equation}
    \begin{split}
 if \  \frac{a}{b} \ < \ \frac{c}{d}\ ,\ &  \frac{a}{b}\ <\  \frac{a+c}{b+d}\ <\ \frac{c}{d}\\
 if \  \frac{\Delta al}{\Delta t_{sd}}< \frac{al(n)}{t_{sd}(n)}\ ,\ &\frac{al(n+1)}{t_{sd}(n+1)} =\frac{al(n)+\Delta al}{t_{sd}(n)+\Delta t_{sd}}<\frac{al(n)}{t_{sd}(n)}
    \end{split}
    \label{eq:sugar}
\end{equation}



\section{Lightweight Sample Reallocation}
\label{sec:allocation}
To address the inefficiencies discussed in Section~\ref{subsec:sample_motivation} and achieve efficient sample reallocation, we aim to attain two goals: (1) determine the reallocation strategy to maximize gains; (2) reduce the migration cost associated with reallocation to prevent negative performance impacts.

\subsection{Reallocation Policy}
\textbf{Target.} For a single instance of speculative decoding, its per-step throughput is defined as follows:
\begin{equation}
    \begin{split}
    Throughput\ = \ \frac{Number \ of \ Accepted\ Tokens}{Execution\ Time}
    \end{split}
    \label{eq:throughput}
\end{equation}

where the execution time is the sum of the draft generation time and the LLM verification time. The \textit{system throughput} should be defined as the sum of the throughput of the individual generation instances, expressed as follows:

\begin{equation}
    \begin{split}
    System\ Throughput\ = \ \sum_{i \in instances} Throughput(i)
    \end{split}
    \label{eq:systhroughput}
\end{equation}

The optimization target of RLHF generation is to process a fixed number of samples in the shortest time, which corresponds to handling a fixed number of tokens. A greater \textit{system throughput} results in a shorter time required to process the fixed tokens. Therefore, the goal of our sample reallocation is to maximize the \textit{system throughput}.

\begin{figure}[htbp]
    \centering
    \includegraphics[width=\columnwidth,trim={0cm 0cm 0cm 0cm}]{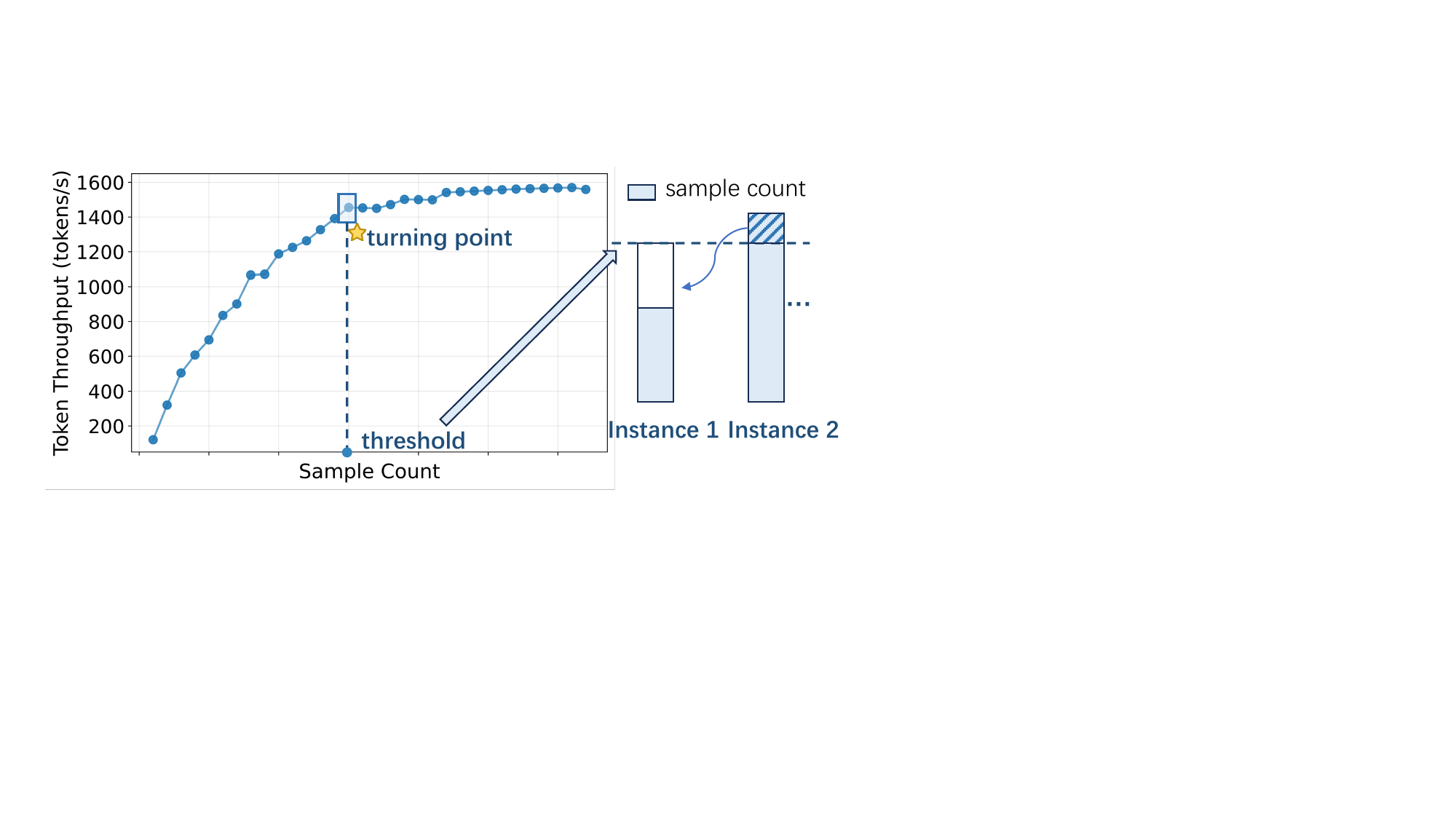}
    \caption{Throughput variation curve of the generation instance over the sample count, and the illustration of the sample reallocation
.}
    \label{fig:sample_throughput}
\end{figure}

\textbf{Observation.} As shown in Figure~\ref{fig:sample_throughput}, we observe that the instance throughput exhibits a roofline phenomenon, initially increasing with the number of samples and then stabilizing. The underlying reason is that when the sample count is limited, an increase in samples leads to a proportional rise in the number of accepted tokens, while the execution time remains relatively constant due to the availability of adequate computational resources, resulting in high marginal returns. However, as the number of samples continues to increase, the execution time also gradually rises, leading to diminishing marginal returns. Thus, when the instance surpasses the sample count corresponding to the turning point (referred to as \textit{threshold}), the benefit of adding one more sample becomes negligible, resulting in minimal performance improvement. Conversely, reallocating samples to instances with fewer samples than the \textit{threshold} will yield greater returns. Therefore, we use a greedy approach that reallocates samples based on the \textit{threshold}. To maximize overall performance, it is essential that as many instances as possible maintain a load that is greater than or equal to the \textit{threshold}. Moreover, the \textit{threshold} is derived from offline profiling data, for which the profiling overhead is negligible in comparison to the RLHF training (details in Section~\ref{subsec:overhead}). We also update the \textit{threshold} based on the data collected during runtime.

\textbf{Reallocation Objectives and Solution.} We consider instances with a sample count below the \textit{threshold} as candidate destination instances (denoted as \textit{d-instance}) for reallocation of samples, while instances with a sample count above the \textit{threshold} are treated as candidate source instances (denoted as \textit{s-instance}). We use $s_{cur}(i)$/$d_{cur}(j)$ and $s_{next}(i)$/$d_{next}(j)$ to denote the sample count of \textit{s-instance} $i$ and \textit{d-instance} $j$ before and after reallocation. The reallocation strategy should meet the following objectives:

\begin{equation}
    \begin{split}
&max{\sum_{j \in \textit{d-instances}} (d_{next}(j)-d_{cur}(j))} \\
s.t. &\forall i \in \textit{s-instances},\ s_{next}(i) >= threshold;\\
&\forall j \in \textit{d-instances},\ d_{next}(j) <= threshold;\\
&\forall k \in {\textit{s-instances},\ \textit{d-instances}},\ m(k)<=1.
    \end{split}
\label{eq:target}
\end{equation}

where $m(k)$ is the number of times instance $k$ triggers migration during a single reallocation decision-making process. The objective is to maximize the number of samples that can be obtained by \textit{d-instances}. There are three constrains: (1) after reallocation, \textit{s-instance} should have a sample count greater than or equal to the \textit{threshold}, as fewer samples may lead to a sudden decline in the \textit{s-instance}'s throughput; (2) after reallocation, \textit{d-instance} should have a sample count less than or equal to the \textit{threshold}, as more samples does not yield addition benefits for \textit{d-instance}'s throughput; (3) an instance will trigger at most one-time sample migration in a single reallocation decision to avoid frequent migration.

We develop a greedy algorithm to solve the problem. We sort the instances based on the sample count in ascending order. Instances with the largest difference will be repeatedly paired until all \textit{d-instances} or all \textit{s-instances} have been matched. For each pair of instances (i,j), the transferred sample count is $min(s_{cur}(i)-threshold, threshold-d_{cur}(j))$. We also need to determine the migration samples. We prefer requests with shorter sequence lengths and a lower average number of accepted tokens. The former is attributed to the fact that a shorter sequence length requires transferring fewer KVCache blocks, resulting in fewer negative performance impacts. The latter is due to the fact that the samples being migrated experience downtime during the migration process. Samples with a lower average number of accepted tokens incur less throughput waste due to downtime, thereby having a smaller impact on overall throughput.

Moreover, to avoid excessive reallocation overhead resulting from frequent sample reallocation, we make the reallocation decision every $cooldown$ steps. A reallocation is triggered if the aforementioned inefficiency is detected.

\subsection{Two-stage Sample Migration}
The significant KVCache states of samples can potentially introduce great cost and generation stalls during reallocation. \textit{RLHFSpec} addresses this challenge by exploiting the key characteristics of speculative decoding: (1) the Markov property of LLM verification; (2) the independence of SSM KVCache and LLM KVCache. For (1), each step of the LLM verification only targets the newly generated speculative tokens, without affecting the results that have already been verified. Thus, for both SSM and LLM, only the KVCache generated in this step will be updated after verification, leaving the KVCache verified by previous steps unchanged. For (2), the KVCache of SSM and LLM are mutually independent, with each computation relying solely on its own KVCache.

Based on the above characteristics, we propose an efficient two-stage migration mechanism.

\begin{figure}[htbp]
    \centering
    \includegraphics[width=0.83\columnwidth,trim={0cm 0.4cm 0cm 0cm}]{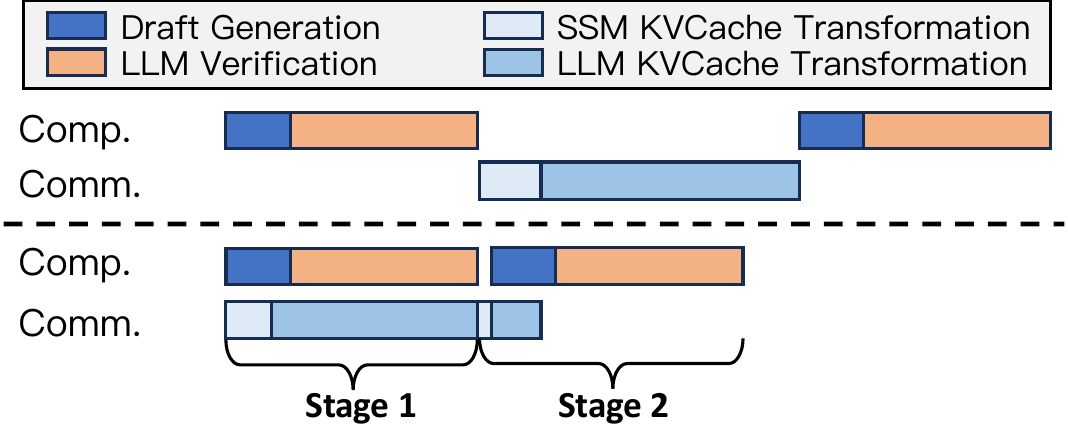}
    \caption{Illustration of two-stage sample migration.}
    \label{fig:migration}
\end{figure}

\textbf{Stage 1: }
We first exploit the Markov property of LLM verification to optimize the migration. Since both the SSM and the LLM's previously verified KV cache will not be modified in the subsequent steps, we can safely transfer the KVCache of previous tokens in parallel with the draft generation and the LLM verification for new tokens. As shown in Figure~\ref{fig:migration}, when the migration is triggered, the \textit{s-instance} begins to transmit the KVCache of verified tokens for both SSM and LLM, while concurrently continuing its computation. Once the KVCache transformation in this stage is complete, only the KVCache of the newly verified tokens that require transmission remains. We further optimize this process in Stage 2.

\textbf{Stage 2: }
We exploit the independence of SSM KVCache and LLM KVCache to overlap the transmission of the remaining KVCache with computation. Since draft generation is solely related to its own KVCache, the \textit{d-instance} can commence the subsequent computation as soon as it receives the SSM KVCache. As shown in Figure~\ref{fig:migration}, once the SSM KVCache transformation is complete, the samples resume on the \textit{d-instance} and initiate the next step of draft generation while the LLM KVCache is concurrently transferring. Upon the completion of the LLM KVCache transmission, the migration is finished, and the LLM performs verification of the newly generated speculative tokens. In this way, we achieve near-zero migration overhead.

The KVCache transmission described above consists of three phases: (1) copying the KVCache from the KVCache store into a buffer; (2) transferring the KVCache from the s-instance buffer to the d-instance buffer; (3) copying the KVCache from the buffer back to the KVCache store.

In phase 1, we transfer multiple layers of KVCache for several samples associated with both SSM and LLM. This involves numerous inefficient copy operations, resulting in non-negligible cudaMalloc and transmission overhead. To reduce this overhead, we define a hierarchical representation for transferring the KVCache. Specifically, we hierarchically organize the K tensor and V tensor according to the order of model (SSM \& LLM)-layer-sample. Based on this representation, we pre-allocate a contiguous space to store all KVCache and transfer them in a single copy operation, effectively reducing cudaMalloc overhead while fully utilizing transmission bandwidth. 

In phase 2, to prevent out-of-memory issues on the \textit{d-instance} after KVCache migration, the \textit{s-instance} will first send an allocation request, specifying the memory required to store all KVCache. Upon receiving this request, the \textit{d-instance} will try to allocate and reserve the memory, informing the \textit{s-instance} of the allocation status. If the allocation is successful, the \textit{s-instance} will proceed to send the KVCache. Otherwise, the \textit{s-instance} will clear the corresponding buffer and report to the reallocator. 

The phase 3 is the reverse process of the phase 1. We parse the received buffer based on the hierarchical representation discussed above, and append the parsed KVCache to the corresponding KVCache stores, ensuring that the migrated samples can be executed correctly on the \textit{d-instance}.

\begin{figure*}[htbp]
    \centering
    \includegraphics[width=0.95\linewidth,trim={0cm 0cm 0cm 0cm}]{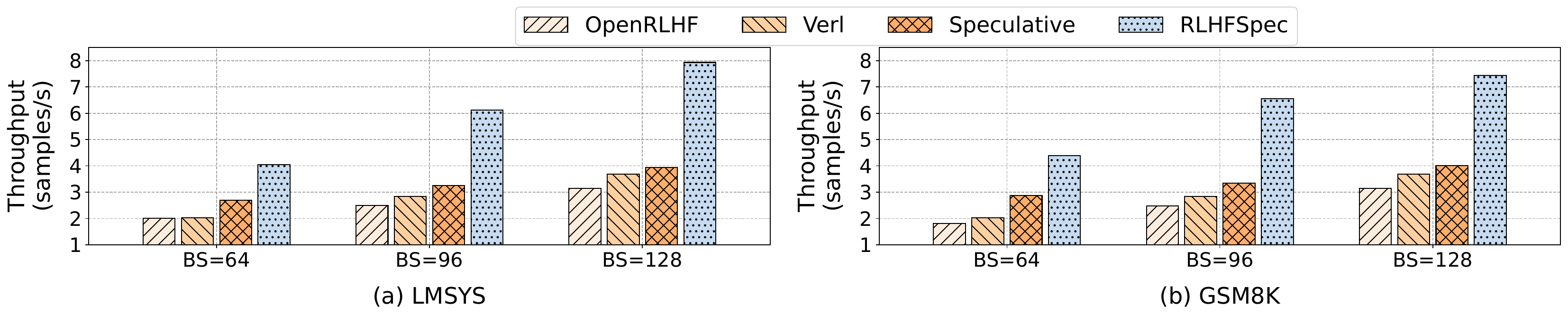}
    \caption{Throughput of different RLHF systems in the generation stage.}
    \label{fig:generation}
\end{figure*}
\begin{figure*}[htbp]
    \centering
    \includegraphics[width=0.97\linewidth,trim={0cm 0cm 0cm 0cm}]{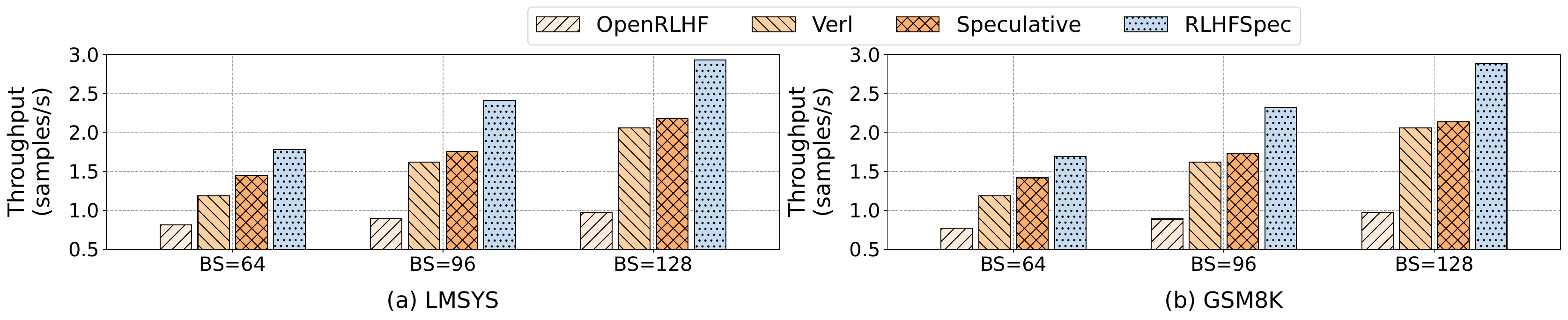}
    \caption{Throughput of different RLHF systems in the entire RLHF execution.}
    \label{fig:e2e}
\end{figure*}
\section{Evaluation}
\label{sec:evaluation}
\subsection{Experiment Setup}
\textbf{Hardware and Software Configurations - }We conduct experiments on a server equipped with 8 NVIDIA L40S GPUs and Intel Xeon Gold 6448Y CPUs. The GPUs are connected over PCIe. The experiments are conducted on Ubuntu 22.04 with CUDA 12.6 and cuDNN v9.1.0. \textit{RLHFSpec} is implemented on \textit{Verl}~\cite{sheng2025hybridflow} (commit: 0759489).

\textbf{Model and Datasets - }We evaluate our system using Llama-3.1-8B-Instruct~\cite{dubey2024llama}, and using corresponding Eagle draft model as SSM~\cite{eagledraft}. Moreover, we utilize the LMSYS-Chat-1M dataset ($LMSYS$) and GSM8K dataset ($GSM8K$) as the training datasets. 

\textbf{Baseline Systems and Comparison Methods - }We compare \textit{RLHFSpec} with state-of-the-art RLHF systems including \textit{Verl}~\cite{sheng2025hybridflow} and \textit{OpenRLHF}~\cite{hu2024openrlhf}. We also compare \textit{RLHFSpec} with \textit{Speculative}, which uses traditional speculative decoding to accelerate RLHF execution. We include this baseline to demonstrate the performance improvements brought by our core techniques in Section~\ref{sec:workload} and Section~\ref{sec:allocation}. For each sample, we allow all systems to generate a maximum of 2048 tokens to avoid out-of-memory issues. We use sample throughput as the evaluation metric, which is defined as the rate of processing samples from start to finish, typically quantified in samples per second (samples/s).

\subsection{RLHF Generation Performance}
\textbf{Throughput -} As shown in Figure~\ref{fig:generation}, on \textit{LMSYS} and \textit{GSM8K}, \textit{RLHFSpec} achieves a maximum speedup of 2.52$\times$/2.65$\times$, 2.16$\times$/2.32$\times$, and 2.02$\times$/1.97$\times$ compared to \textit{OpenRLHF}, \textit{Verl}, and \textit{Speculative}, respectively. Note that compared with our base system \textit{Verl}, on \textit{LMSYS} and \textit{GSM8K}, \textit{RLHFSpec} achieves an average speedup of 2.10$\times$ and 2.17$\times$ respectively, which further demonstrates the advantages of our work. \textit{Speculative} exhibits performance enhancements in comparison to \textit{OpenRLHF} and \textit{Verl}, indicating that it can partially mitigate inefficiencies during the generation process. However, the performance gap compared to \textit{RLHFSpec} indicates that directly applying speculative decoding to RLHF does not fully exploit its potential.

\textbf{Understanding the Performance -} \textit{RLHFSpec} achieves better performance for the following reasons. Compared to existing RLHF systems with autoregressive decoding, \textit{RLHFSpec} can utilize spare computational resources to perform efficient speculative decoding with a high acceptance rate of tokens while maintaining the same per-step latency as autoregressive decoding. Compared to existing speculative decoding work, \textit{RLHFSpec} dynamically selects the near-optimal drafting strategy for varying workloads, greatly balancing the verification cost and the number of accepted tokens. Moreover, \textit{RLHFSpec} identifies the limitation of fixed sample reallocation in achieving high system throughput and proposes an efficient sample reallocation strategy to fully utilize the computational resources on spare instances. We provide a detailed illustration of the aforementioned improvements in Section~\ref{subsec:ablation}, Section~\ref{subsec:optimal}, and Section~\ref{subsec:deepdive}.

\subsection{End-to-end Performance}
We also evaluate the performance of different systems in the entire RLHF execution. As shown in Figure~\ref{fig:e2e}, on \textit{LMSYS} and \textit{GSM8K}, \textit{RLHFSpec} achieves a maximum speedup of 3.01$\times$/2.97$\times$, 1.50$\times$/1.43$\times$, and 1.37$\times$/1.35$\times$ compared to \textit{OpenRLHF}, \textit{Verl}, and \textit{Speculative}, respectively. Note that compared with our base system \textit{Verl}, on \textit{LMSYS} and \textit{GSM8K}, \textit {RLHFSpec} achieves an average speedup of 1.47$\times$ and 1.42$\times$ respectively, which further demonstrates the advantages of our work. Moreover, the low throughput of \textit{OpenRLHF} results from the lack of support for model parameter offloading, which necessitates the retention of multiple sets of model parameters. To mitigate out-of-memory issues, \textit{OpenRLHF} uses smaller micro batch sizes during the training stage, which leads to a substantial increase in training time. The superior performance of \textit{RLHFSpec} demonstrates that by alleviating generation bottlenecks, \textit{RLHFSpec} has significantly improved the overall performance of RLHF.

\subsection{Performance Breakdown}
\label{subsec:ablation}
\begin{figure}[htbp]
    \centering
    \includegraphics[width=0.96\columnwidth,trim={0cm 0cm 0cm 0cm}]{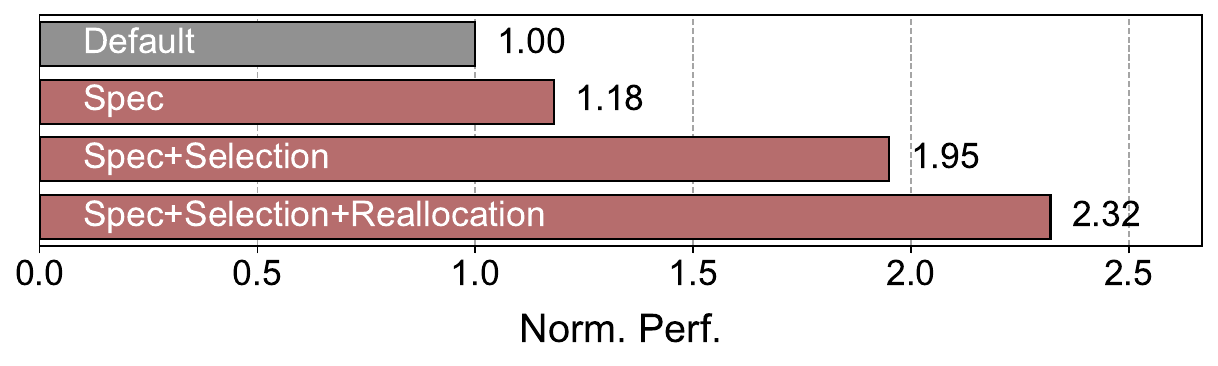}
    \caption{Throughput breakdown of \textit{RLHFSpec}. Reported throughputs are normalized by \textit{Default}.}
    \label{fig:breakdown}
\end{figure}

To better understand the contribution of adopted methods in \textit{RLHFSpec}, we profile and break down the generation performance of \textit{RLHFSpec} into three parts. We refer to the autoregressive decoding as the baseline (\textit{Default}). We first employ speculative decoding ($Spec$) on the basis of \textit{Default}. As shown in Figure~\ref{fig:breakdown}, \textit{Spec} can increase the throughput by 1.18$\times$. This is because \textit{Spec} addresses the limited degree of parallelism in autoregressive decoding and leverages the spare GPU resources during generation execution.

On top of \textit{Spec}, we further evaluate the effectiveness of the workload-aware drafting strategy selection (\textit{Selection}). These optimizations together improve the throughput by 1.95$\times$. This is because \textit{Selection} can dynamically adjust the drafting strategy to bring it closer to the optimal, greatly balancing the verification cost and the number of accepted tokens under varying workloads.

On top of \textit{Spec} and \textit{Selection}, we further evaluate the effectiveness of sample reallocation (\textit{Reallocation}). These optimizations together improve the throughput by 2.32$\times$. This is because \textit{Reallocation} effectively harnesses idle GPU resources on spare instances to achieve high system throughput while incurring minimal reallocation overhead.

\subsection{Effectiveness of Workload-aware Drafting Strategy Selection}
\label{subsec:optimal}
To illustrate the effectiveness of our workload-aware drafting strategy selection, we have evaluated the implementations with fixed \textit{draft token num} $n$ across varying workloads (i.e., sample counts). We found that when $n$ exceeds 48, performance inevitably drops due to high verification cost. Thereby, our comparison uses $n$ ranging from 2 to 48, consistent with the range used in existing studies~\cite{miao2024specinfer,li2025adaserve}. We regard the highest performance across all configurations as \textit{optimal}. As shown in Table~\ref{table:strategy_exp}, across various datasets and workloads, \textit{RLHFSpec} consistently achieves high throughput that closely approximates the \textit{optimal}. Furthermore, even in the worst-case scenario, \textit{RLHFSpec} is able to attain 95.53\% of the \textit{optimal}. The experimental results demonstrate that \textit{RLHFSpec} is capable of accurately selecting near-optimal drafting strategies across different workloads, thereby fully leveraging the performance enhancements offered by speculative decoding under the dynamically varying generation workloads.

\begin{table}[h]
    \centering
    \caption{Throughput of \textit{RLHFSpec} compared to the \textit{optimal} under different workloads, where the \textit{optimal} indicates the highest throughput across all drafting strategies.}
    \begin{tabular}{|l|c|c|}
    \hline
    \textbf{Workloads} & \textbf{LMSYS Dataset} & \textbf{GSM8K Dataset}  \\ 
    \hline
    sample count = 8   & 96.57\%   & 96.99\%      \\
    \hline
    sample count = 16   & 97.33\%   & 98.30\%      \\
    \hline
    sample count = 24   & 97.76\%   & 95.53\%     \\
    \hline
    sample count = 32 & 98.27\%   & 99.90\%     \\
    \hline
    sample count = 40 & 99.70\%   & 99.76\%     \\
    \hline
    sample count = 48 & 98.89\%   & 98.00\%     \\
    \hline
    sample count = 56 & 98.90\%   & 98.72\%     \\
    \hline
    sample count = 64 & 97.97\%   & 99.67\%     \\
    \hline
    
    \end{tabular}
    \label{table:strategy_exp}
\end{table}

\subsection{Deep Dive into the Sample Reallocation}
\label{subsec:deepdive}
\begin{figure}[htbp]
    \centering
    \includegraphics[width=\columnwidth,trim={0cm 0cm 0cm 0cm}]{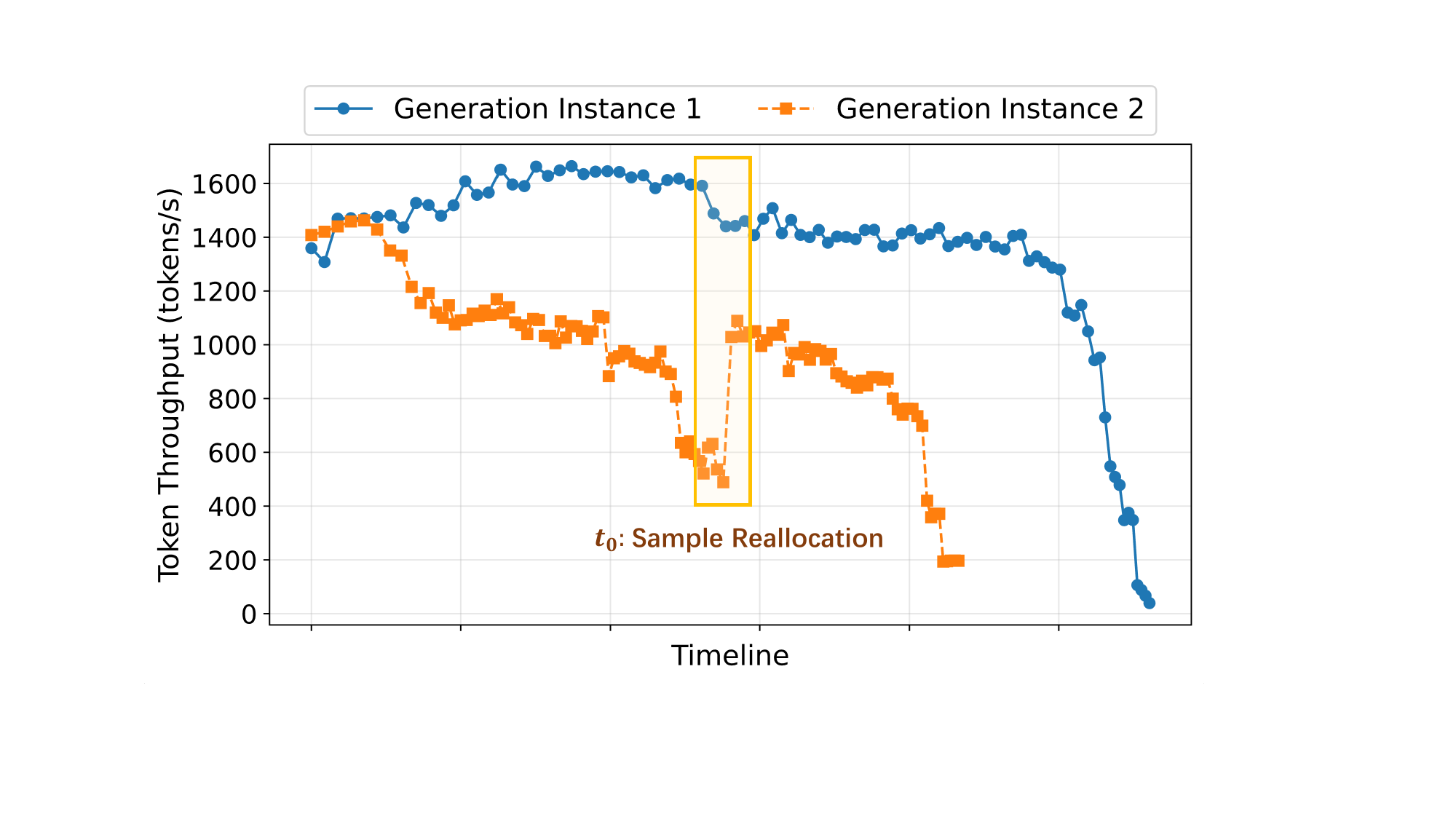}
    \caption{Throughput variation curves of two generation instances over time.}
    \label{fig:throughput}
\end{figure}

We deep dive into the sample reallocation by presenting the throughput variation curves of two generation instances. As shown in Figure~\ref{fig:throughput}, \textit{instance 2} is allocated a greater number of short samples, resulting in a more rapid decline in sample count compared to \textit{instance 1}. At time $t_0$, the reallocator monitors that \textit{instance 2}, due to containing only a limited number of samples, is unable to fully utilize GPU resources. Consequently, the reallocator decides to migrate five samples from \textit{instance 1} to \textit{instance 2}. Following this migration, the system's token throughput increases from 2,127 tokens/s to 2,531 tokens/s, resulting in a significant performance enhancement and demonstrating the effectiveness of our proposed sample reallocation.

\subsection{Overhead Analysis}
\label{subsec:overhead}
The overhead introduced by \textit{RLHFSpec} comes from three parts, including workload-aware drafting strategy selection ($WDS$), sample reallocation decision ($SRD$), and sample migration ($SM$). $WDS$ predicts the number of accepted tokens and execution time of the tokens in the speculative tree, and traverses the tree to select the near-optimal strategy. $SRD$ determines the reallocation policy based on the workload of each instance. $SM$ migrates the KVCache of samples across instances. The overhead incurred by $WDS$, $SRD$, and $SM$ is less than 3.87\% of the total execution time, which imposes little impact on generation execution. There are also offline overhead associated with data collection to train the predictors and determine the throughput turning points. This overhead totals approximately 15 minutes and is a one-time cost, which is acceptable in comparison to the longer duration of RLHF training.
\section{Related Work}

\subsection{RLHF Training Optimizations}
Researchers have proposed various optimizations to accelerate RLHF training~\cite{sheng2025hybridflow, hu2024openrlhf,mei2024realhf,zhong2025streamrl,yao2023deepspeed,havrilla2023trlx, xiao2023adaptive}. Verl~\cite{sheng2025hybridflow} proposes a hierarchical hybrid programming model to enable flexible dataflow representation and efficient execution. OpenRLHF~\cite{hu2024openrlhf} places each model on separate devices and designs customized parallel strategies for different tasks. ReaLHF~\cite{mei2024realhf} formulates the RLHF execution as an augmented dataflow and uses a search algorithm to discover efficient execution plans. 
However, these works fundamentally employ autoregressive decoding during the generation stage, leading to inefficient execution and low resource utilization. In contrast, \textit{RLHFSpec} utilizes the performance potential derived from the application of speculative decoding in generation execution and proposes an workload-aware drafting strategy selection mechanism, along with sample reallocation, to fully exploit these opportunities. Through these methods, \textit{RLHFSpec} overcomes the constraints of autoregressive decoding, which is orthogonal to existing RLHF works.

\subsection{Speculative Decoding}
Speculative decoding~\cite{miao2024specinfer,li2024eagle,butler2024pipeinfer,leviathan2023fast,chen2023accelerating,cai2024medusa,zhao2024ouroboros} serves as an efficient method to improve the process of LLM autoregressive decoding. SpecInfer~\cite{miao2024specinfer} combines multiple boost-tuned SSMs and builds a speculative tree to predict the output of the LLM. Eagle~\cite{li2024eagle} constructs the speculative tree based on the context and selects the top-k tokens with the highest inference logits for verification. However, these methods mainly focus on online serving scenarios. RLHF generation resembles offline inference, where the total number of processing samples is fixed and the optimization goal is to complete all samples in the shortest time. \textit{RLHFSpec} efficiently addresses the additional dynamic load and load imbalance introduced by RLHF generation, and achieves significant speedup.  

\section{Conclusion}
In this paper, we introduce \textit{RLHFSpec}, an RLHF system featuring efficient speculative decoding and sample reallocation. \textit{RLHFSpec} designs a workload-aware drafting strategy that dynamically selects the drafting strategy for varying generation workloads. \textit{RLHFSpec} also proposes sample reallocation as the key measure to fully utilize the GPU resources and thus improve the overall performance. To reduce the sample reallocation cost, an efficient two-stage sample migration is employed. Evaluation results show that \textit{RLHFSpec} delivers higher throughput in RLHF execution. 

\bibliographystyle{ACM-Reference-Format}
\bibliography{references}
\clearpage
\end{document}